\documentclass[10pt,twocolumn,letterpaper]{article}

\usepackage{cvpr}              

\usepackage{graphicx}
\usepackage{amsmath}
\usepackage{amssymb}
\usepackage{booktabs}
\usepackage{animate}
\usepackage[export]{adjustbox}
\usepackage{microtype}
\usepackage[dvipsnames]{xcolor}
\usepackage{eucal}
\usepackage{multirow}

\newcommand{\eqnref}[1]{Eq.~(\ref{#1})}

\usepackage[pagebackref,breaklinks,colorlinks]{hyperref}
\usepackage{mwe}

\usepackage[capitalize]{cleveref}
\crefname{section}{Sec.}{Secs.}
\Crefname{section}{Section}{Sections}
\Crefname{table}{Table}{Tables}
\crefname{table}{Tab.}{Tabs.}

\def\cvprPaperID{4632} 
\def\confName{CVPR}
\def\confYear{2023}

\title{High-fidelity 3D Human Digitization from Single 2K Resolution Images}

\author{Sang-Hun Han$^{1}$, Min-Gyu Park$^{2}$, Ju Hong Yoon$^{2}$, \\Ju-Mi Kang$^{2}$, Young-Jae Park$^{1}$ and Hae-Gon Jeon$^{1}$\\
$^{1}$Gwangju Institute of Science and Technology (GIST), $^{2}$Korea Electronics Technology Institute (KETI)\\
{\tt\small \{sanghunhan, youngjae.park\}@gm.gist.ac.kr, haegonj@gist.ac.kr}\\
{\tt\small \{mpark, jhyoon, yypeip\}@keti.re.kr}
}

\begin{document}

\twocolumn[{%
\renewcommand\twocolumn[1][]{#1}%
\maketitle
\maketitle
\begin{center}
    \vspace{-18pt}
    \centering
    \captionsetup{type=figure}
    \includegraphics[width=2.0\columnwidth]{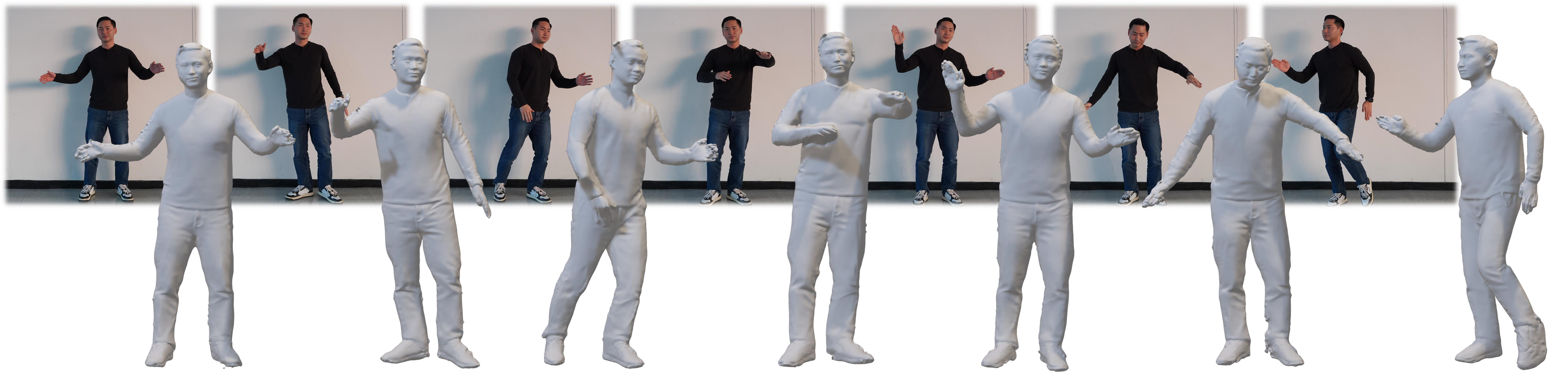}
    \vspace{-7pt}
    \captionof{figure}{Frame-by-frame reconstruction results for a 2K resolution image sequence. The image sequence is captured by a smartphone.}
    \label{fig:intro}
\end{center}%
}]

\begin{abstract}\vspace{-2mm}
High-quality 3D human body reconstruction requires high-fidelity and large-scale training data and appropriate network design that effectively exploits the high-resolution input images. To tackle these problems, we propose a simple yet effective 3D human digitization method called \textbf{2K2K}, which constructs a large-scale 2K human dataset and infers 3D human models from 2K resolution images. The proposed method separately recovers the global shape of a human and its details. The low-resolution depth network predicts the global structure from a low-resolution image, and the part-wise image-to-normal network predicts the details of the 3D human body structure. The high-resolution depth network merges the global 3D shape and the detailed structures to infer the high-resolution front and back side depth maps. Finally, an off-the-shelf mesh generator reconstructs the full 3D human model, which are available at \url{https://github.com/SangHunHan92/2K2K}. In addition, we also provide 2,050 3D human models, including texture maps, 3D joints, and SMPL parameters for research purposes. In experiments, we demonstrate competitive performance over the recent works on various datasets.
\end{abstract}\vspace{-2mm}

\section{Introduction}
\label{sec:intro}
Reconstructing photo-realistic 3D human models is one of the actively researched topics in computer vision and graphics. Conventional approaches search for correspondences across multiple views. Therefore, it was necessary to employ multiple camera systems~\cite{Collet_2015_TOG,dou2016fusion4d} to acquire high-quality human models. However, the bulky and expensive camera systems limit the usage of normal users such as personal content creators and influencers. Recent progress in deep learning has shown the possibility of reconstructing human models from a single image~\cite{Varol_2018_ECCV,Saito_2019_ICCV,Zheng_2019_ICCV,Zheng_2021_PaMIR,alldieck2019learning,kanazawa2018end,natsume2019siclope, He_2020_NeurIPS,pesavento2022super,xiu2022icon}. Nevertheless, there still exists room to improve the quality of 3D human models, especially given an input single image.

Existing approaches fall into two categories; the first is to predict a deep implicit volume~\cite{Saito_2019_ICCV,Saito_2020_CVPR,He_2020_NeurIPS} and the second is to infer multiple depth maps~\cite{Gabeur_2019_ICCV} from an image. In the case of the first approach, the implicit volume can be directly predicted through a deep learning network~\cite{Varol_2018_ECCV} or the volume can be constructed by predicting each voxel~\cite{Saito_2019_ICCV,Saito_2020_CVPR} millions of times. Therefore, these approaches are demanding either in memory or in time. The second approach requires to predict at least two depth maps, one for the front and the other for the back, to build a complete 3D human model. Gabeur \etal~\cite{Gabeur_2019_ICCV} propose an adversarial framework to predict double-sided depth maps; however, it shows poor results owing to inadequate training data generated by using 19 human scan models in addition to the synthetic dataset~\cite{Varol_2017_CVPR}. Here, researchers have paid attention to recover not only geometric details~\cite{Saito_2020_CVPR} such as facial regions but also people in various postures by predicting surface normal maps and parametric template models~\cite{SMPL-X_2019_Graph}. 

We claim that the prediction quality of 3D human models is primarily affected by suitably designed deep neural networks and high-quality training datasets, \ie, high-resolution images and 3D human models. Existing approaches, however, could not handle high-resolution images,   \eg, 2048$\times$2048, because it requires massive learnable parameters to train, and there was no such large-scale dataset for 3D human digitization.

There are human datasets opened publicly for research purposes~\cite{Zhang_2017_BUFF,yu2020humbi,Zheng_2019_ICCV,tao2021function4d}. These datasets, however, lack both quality and quantity to train a network generating high-fidelity human models. 
In this paper, we present a practical approach to reconstructing high-quality human models from high-resolution images and a large-scale human scan dataset consisting of more than 2,000 human models, where existing methods utilize a few hundred human scans to train their networks.

Our\,framework\,takes\,high-resolution\,images\,as\,input up to 2K, 2048$\times$2048, and it\,is\,the\,first\,to\,predict high-resolution depth maps for the task of 3D human reconstruction, named \textbf{2K2K}. To minimize the number of learnable parameters and memory usage, we split the human body into multiple body parts such as arms, legs, feet, head, and torso, with the aid of a 2D pose human detector~\cite{cao2017realtime}.\,In addition, we align each body part by rotating and scaling to a canonical position which makes the proposed method robust under human pose variation while excluding background regions from the computation. By doing this, the part-wise image-to-normal prediction network can predict accurate surface normals even in the presence of pose variations. Afterward, we merge predicted normal maps into a single normal map and feed it to the normal-to-depth prediction network. Note that it is hard to predict depth maps directly for each body part because of the scale ambiguity; predicting the depth map from a merged normal map can alleviate this problem.\,We also predict a coarse depth map and feed it to the normal-to-depth prediction network to obtain consistent depth maps over different body parts. Finally, we generate high-fidelity human meshes through Marching cubes~\cite{Lorensen_1987_SIG}, whose example is shown in Fig.~\ref{fig:intro}.

To summarize, the contributions of this paper are:
\begin{enumerate}
    \item \textbf{Accuracy.} Our method recovers the details of a human from high-resolution images up to a resolution of 2048$\times$2048. \vspace{-3mm}
    \item \textbf{Efficiency.} The part-wise normal prediction scheme naturally excludes background regions from computation, which results in a reduced memory footprint.\vspace{-3mm}
    \item \textbf{Data.} We release a large-scale 3D human dataset consisting\,of\,2,050\,human models with synthesized images.\vspace{-3mm}
\end{enumerate}

\begin{figure*}[t]
\centering
\includegraphics[width=2.0\columnwidth]{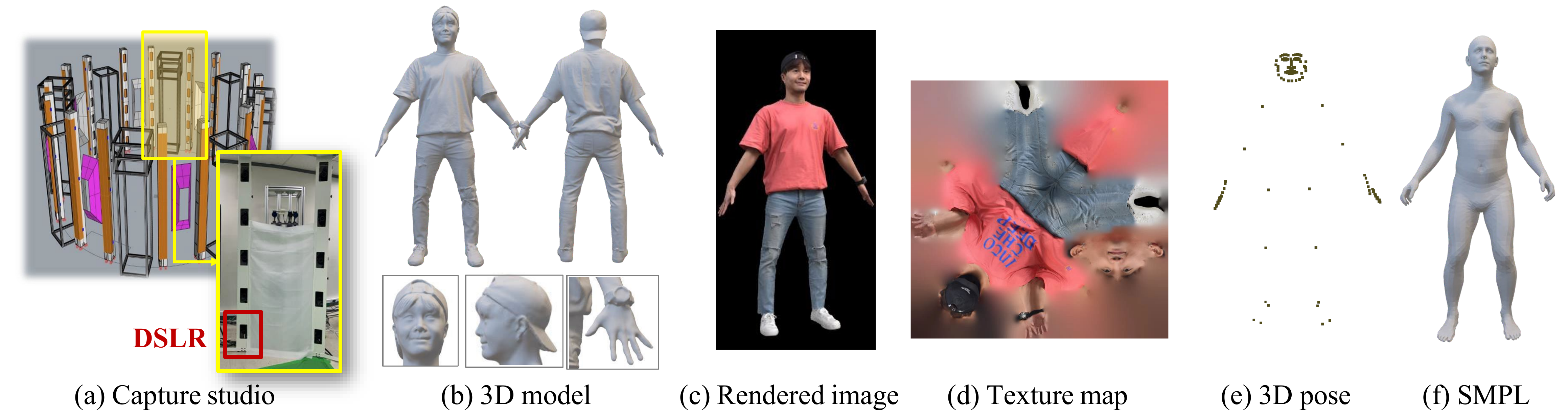} 
\vspace{-3.5mm}
\caption{Our datasets provides high-fidelity 3D human models, captured by 80 DSLR cameras, texture maps, 3D poses (openpifpaf-wholebody \cite{kreiss2021openpifpaf}), and SMPL model parameters \cite{SMPL_2015_Graph}.}
\vspace{-2.5mm}
\label{fig:studio}
\end{figure*}

\begin{figure*}[t]
\centering
\includegraphics[width=2.1\columnwidth]{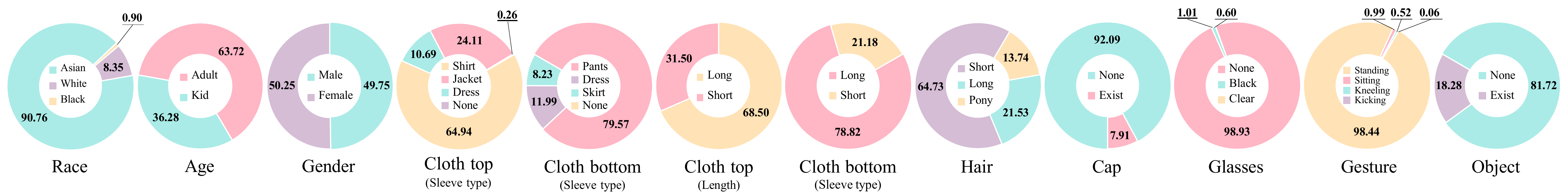}
\vspace{-5.5mm}
\caption{Statistics of the  scanned human models in the proposed dataset.}
\vspace{-4mm}
\label{fig:statistics}
\end{figure*}

\section{Related Work}
\label{sec:related}

We review clothed human 3D reconstruction \textcolor{black}{from single images with respect to datasets}, model-free approaches, and parametric model-based approaches. Here, we refer to the parametric model as the Skinned Multi-Person Linear (SMPL) model~\cite{SMPL_2015_Graph, SMPL-X_2019_Graph} that has been widely used for human pose estimation and reconstruction.

\begin{table}[t]
\setlength{\tabcolsep}{3pt}
\centering
\begin{center}
\renewcommand{\arraystretch}{1.2} 
    \resizebox{\linewidth}{!}{%
	\begin{tabular}{c| c | c | c | c | c | c | c | c | c}
		\noalign{\hrule height 1pt}   
		 Dataset & License & \# of subjects & \# of Frames & \# or vertices & RGB & Mesh & Texture & SMPL(-X) & Keypoint3d \\ 
		 \hline
		 \hline
		 BUFF~\cite{Zhang_2017_BUFF} & Non-commercial & 6    & $>$13.6k & ~150k & \checkmark & \checkmark & \checkmark & \checkmark & \checkmark \\ 
		 CAPE~\cite{CAPE:CVPR:20}    & Non-commercial & 15   & $>$140k  & 6890 &  & \checkmark &  & \checkmark & \checkmark\\ 
		 DFAUST~\cite{dfaust:CVPR:2017}       & Non-commercial & 10   & $>$40k   & 190k & \checkmark & \checkmark & \checkmark & \checkmark & \checkmark \\ 
		 Humbi~\cite{yu2020humbi}        & Non-commercial & 772  & ~26M     & 50k & \checkmark & \checkmark & \checkmark & \checkmark & \checkmark\\ 
		 THuman~\cite{Zheng_2019_ICCV}       & Non-commercial & 200  & 6000      & 150k & \checkmark & \checkmark & \checkmark & \checkmark & \\ 
		 THuman2.0~\cite{tao2021function4d}    & Non-commercial & 200  & 526      & 300k &     & \checkmark & \checkmark & \checkmark & \\
		 MultiHuman~\cite{zheng2021deepmulticap}   & Non-commercial & 150   & 453      & ~500k & \checkmark & \checkmark &  & \checkmark & \\ 
		 HuMMan~\cite{cai2022humman}       & Non-commercial & 1000 & 60M      & ~300k & \checkmark & \checkmark & \checkmark & \checkmark & \checkmark\\ 
    		 RenderPeople~\cite{Renderpeople} & Commercial     & -    & -        & ~223k & \checkmark & \checkmark & \checkmark &  & \\ 
		 AXYZ~\cite{AXYZ}         & Commercial     & -    & -        & ~100k & \checkmark & \checkmark & \checkmark &   & \\ 
		 Twindom~\cite{Twindom}      & Commercial     & -    & -        & ~200k & \checkmark & \checkmark & \checkmark &  & \\ 
		 \hline
		 \hline
		 2K2K (Ours)         & Non-commercial & 2050 & 2050   & 1M & \checkmark & \checkmark & \checkmark & \checkmark & \checkmark\\ 
         \hline
         \noalign{\hrule height 1pt} 
	\end{tabular}%
	}
\end{center}
\vspace{-15pt}
\caption{A summary of the clothed human body dataset.}
\label{table:dataset}\vspace{-10pt}
\end{table}

\subsection{Datasets for dressed human reconstruction}
The clothed human dataset is fundamental for the study of reconstructing the 3D human surface. 
Early studies use a limited amount of data~\cite{Zhang_2017_BUFF, dfaust:CVPR:2017} consisting of few subjects or simple posture.
Subsequently, as the parametric model gain popularity in this field, a dataset~\cite{CAPE:CVPR:20} expressing cloth by modifying SMPL vertices appeared.
However, it become necessary to reconstruct human surfaces with more complex postures and difficult clothes. Therefore, a number of studies~\cite{Zheng_2019_ICCV, tao2021function4d, zheng2021deepmulticap} capture their own datasets and release the data to the public.  

On the other hand, commercial human datasets~\cite{Renderpeople, AXYZ, Twindom} are used for research to ensure professional quality. However, these datasets cost a lot for research.
Recently, large-scale human datasets~\cite{yu2020humbi, cai2022humman} are released but they lack geometry information for detailed face and cloth reconstruction.
To overcome this, we release a non-commercial 3D human dataset with the largest number of subjects, \ie humans, and high-resolution geometry features.
The characteristics of each dataset are summarized in Table.~\ref{table:dataset}.
		 
\subsection{Model-free human reconstruction}
BodyNet~\cite{Varol_2018_ECCV} and DeepHuman~\cite{Zheng_2019_ICCV} exploit human body segmentation or 2D poses and combine them to predict deep implicit volumes. However, the end-to-end training of a volume prediction network requires a large memory footprint and computation. For this reason, BodyNet represents the 3D shape of the human body with a voxel volume whose resolution is $128\times128\times128$. \textcolor{black}{Caliskan \etal~\cite{caliskan2021temporal} use an additional implicit 3D reconstruction module to overcome the low resolution.}
As one of the model-free approaches, depth prediction~\cite{jinka2020peeledhuman, tang2019neural} is an effective approach to building 3D human models. For example, Mustafa \etal~\cite{mustafa2021multi} show that multiple humans can be reconstructed with the aid of instance segmentation. 
ARCH~\cite{huang2020arch} and ARCH++~\cite{he2021arch++} recover the body geometry in a canonical space not only to predict accurate human models but also to animate reconstructed human models. 
PIFu~\cite{Saito_2019_ICCV} and its variants \cite{He_2020_NeurIPS,hong2021stereopifu} train a deep implicit function defined over a pre-defined space. 
Once the implicit function is trained, it is possible to construct an implicit volume by using the implicit function repeatedly, \ie, for each voxel.
PHORHUM~\cite{alldieck2022photorealistic} predicts both geometric and photometric information simultaneously by using global illumination and surface normal prediction.
PIFuHD~\cite{Saito_2020_CVPR} presents the implicit function for high-resolution images whose input size is up to $1024\times1024$.
They generate high-and low-resolution feature maps and embed them to compute the 3D occupancy field for reconstructing a high-fidelity 3D human model accurately. 
However, PIFuHD still needs much memory and has slow inference time due to the per-pixel calculation to construct coarse and fine implicit volumes.

Our body part-wise approach allows fast, lightweight, and high-fidelity 3D reconstruction for human models against existing model-free approaches. In addition, we alleviate the potential problems of the model-free approach such as robustness under pose variations. 

\subsection{Parametric model-based human reconstruction}
Pavlakos \etal~\cite{SMPL-X_2019_Graph} introduce SMPL-X, which proposes latent pose features to exclude invalid human poses efficiently.
SMPLpix~\cite{prokudin2021smplpix} estimates realistic images with various viewpoints and poses from sparse 3D vertices using RGB-D images obtained from the SMPL model.
PaMIR~\cite{Zheng_2021_PaMIR} fuses the SMPL model features and image features in the implicit field to reconstruct the detailed 3D human model. Moreover, thanks to the initial geometry cue from the SMPL, it can robustly deal with depth ambiguity. 
Tan \etal~\cite{tan2020self} propose a supervised approach to predict depth maps from a video by imposing the consistency between different frames through the initial SMPL model.
SMPLicit~\cite{corona2021smplicit} predicts the unsigned distance to the object surface using the SMPL occlusion map in the latent space.
Zhu \etal \cite{zhu2021detailed,zhu2019detailed} propose to deform initial SMPL vertices using image-aligned features.
ICON~\cite{xiu2022icon} reconstructs clothed human surfaces in various postures by feeding initial normal maps from the predicted SMPL model.
DensePose~\cite{guler2018densepose} introduces a simplified human mesh model that is represented by the predefined UV map. 
Alldieck \etal \cite{Alldieck_2019_CoRR} utilize DensePose to extract and project human image textures in the UV map. Their method first infers surface normal maps from the UV texture map and afterward reconstructs a 3D human model in a single predefined pose, \eg, T or A pose.
Jafarian and Park~\cite{Jafarian_2021_CVPR} adopt DensePose to infer depth information from the video. Having the depth information of a previous frame in one hand, they sequentially improve the details of 3D human body structures. 
Zhao \etal \cite{zhao2022high} modify the SMPL templates for A-posed humans and compute the dynamic offset through the UV map to reconstruct human models. 

Unlike the above works, our 2K2K only needs single images to capture fine 3D details. In addition, the proposed method can deal with various human poses, clothes, and accessories because it does not count on the parametric human model, which has difficulty capturing geometric details.

\begin{figure*}[t]
\centering
\includegraphics[width=1.0\linewidth]{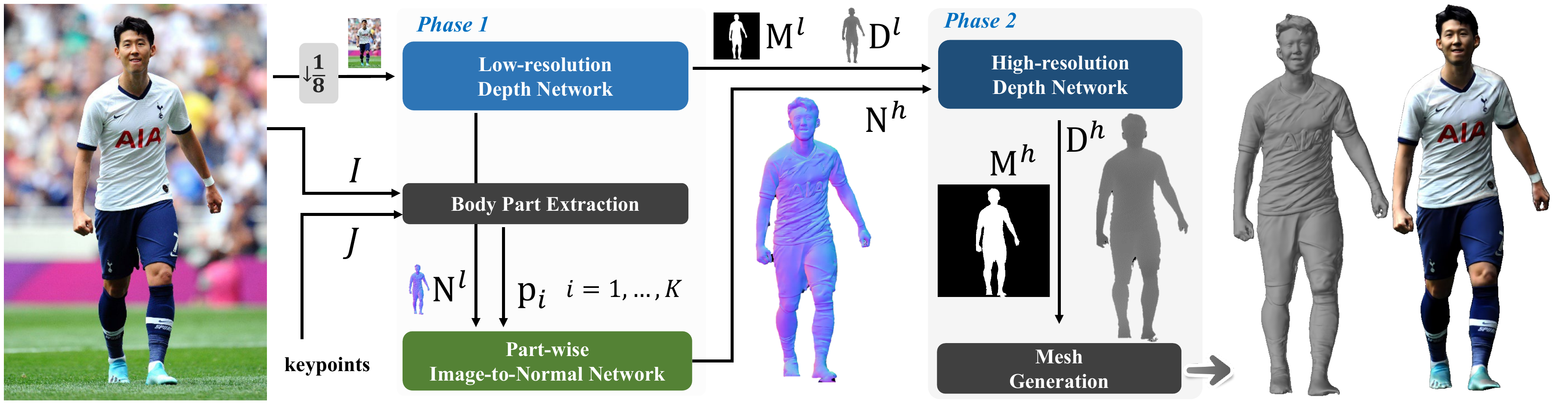}
\vspace{-20pt}
\caption{An overall framework of the proposed method. The first phase predicts the low-resolution front/back view depth maps, and the high-resolution front/back view normal maps. The high-resolution depth network upsamples the low-resolution depth maps with the guidance of the high-resolution normal maps. Finally, the mesh generation reconstructs the full 3D model.}
\vspace{-8pt}
\label{fig:overall_framework}
\end{figure*}

\section{Human Scan Dataset}
\label{sec:dataset}%
Since the performance of human reconstruction primarily depends on the quality and quantity of training data, we release our high-quality human scan models in public, consisting of 2,050 3D human models scanned by our scan booth. The booth contains 80 DSLR cameras, \ie, Nikon D50 model, where the cameras are aligned vertically and horizontally as illustrated in Fig.~\ref{fig:studio} (a). The pitch angle of a camera is adjusted manually, assuming an average human height is standing in the middle, \ie, 5.9 feet tall. The highest cameras look at the head region; the second-highest cameras look at the upper body; the third-highest cameras capture the mid-body region; the fourth-highest cameras look at the lower body; the lowest cameras are pointing at the knees. We generate initial human models by using commercial software, \ie, RealityCapture~\cite{reality}, and manually post-process the models by filling holes and enhancing geometric details in hair regions by experts. Finally, we decimate meshes while keeping important geometric structures. The number of vertices for released models is about 1M, whose example is shown in Fig.~\ref{fig:studio} (b). The scanned model preserves the geometric details such as fingers and wrinkles properly mainly because of high-quality images captured in a controlled environment.

\subsection{Synthesizing images for training}%
Instead of directly using images taken in our scan booth, we augment our data to mimic realistic environments and diversify the training data. There are many ways to render such images~\cite{Varol_2017_CVPR}. In terms of synthesizing approaches, it is possible to generate photo-realistic images with environmental assets and rendering software such as Unreal~\cite{unrealengine} and Unity~\cite{Unity}. An example result is shown in Fig.~\ref{fig:studio}(c), where the reflecting materials and shadings are rendered realistically. However, this ray-tracing approach is time-consuming and requires manual settings by experts. Therefore, we use a more straightforward approach \ie, the Phong lighting model while superimposing the background using randomly selected images. 
Therefore, one can also recover shadings and lighting directions from the shaded input. 
In addition, we also note that multiple humans can be placed at random positions, or a single person can be placed at the center of the image. 
The latter assumes that humans are already detected and appropriately cropped, thanks to the recent advances in object detection technology. 
We additionally provide texture map, 3D keypoint, and SMPL Model as shown in Fig.~\ref{fig:studio}(d),(e),(f) for the effectiveness of the dataset.
In this paper, we generate human-centered training images, whose details will be described in Sec.\ref{sec:exp}.

\subsection{Statistics and bias}
There are many kinds of human appearances in the real world, determined by various factors such as gender, country, and culture. 
We believe, therefore, it is crucial to share statistics of scanned humans to make our dataset applicable to diverse research topics. In order to compile distribution statistics for our dataset, we limit those to 12 categories as shown in Fig.~\ref{fig:statistics}. Due to the geographic circumstances, more than 90\% of models are from Asian countries, which is a common issue on the person-related dataset. By considering various combinations of the treat, our dataset enables deep learning networks to form diverse feature representations of human bodies.

\section{Proposed Method}
\label{sec:proposed}

We aim to produce a high-quality 3D human model $\mathbf{V}$ from a single high-resolution image $I$ with a resolution of up to $2048\times2048$. As illustrated in Fig. \ref{fig:overall_framework}, the proposed method $\Phi(\cdot)$ conducts reconstruction formulated as below:
\begin{equation}
    \begin{array}{ll}  
         \mathbf{V} & = \Phi (I, \mathcal{J}, \mathcal{S}),
    \end{array}
    \label{eq:proposed_method}
\end{equation}
where $j_i \in \mathcal{J}$ refers an $i^\mathrm{th}$ joint and $s_i \in \mathcal{S}$ denotes predefined patch size cropped centered at $j_i$. Using these inputs, the first phase predicts low-resolution depth maps $\mathbf{D}^l$ and high-resolution normal maps $\mathbf{N}^h$ \textcolor{black}{from low-resolution normal maps $\mathbf{N}^l$}. The second phase produces the high-resolution front and back-sided depth maps $\mathbf{D}^h$ from $\mathbf{D}^l$ and $\mathbf{N}^h$. Finally, the high-fidelity 3D model $\mathbf{V}$ is reconstructed from $\mathbf{D}^h$. 

\subsection{Part-wise image-to-normal network} 
\label{sec:partwise} 

\begin{figure}[t]
\centering
\includegraphics[width=1.0\linewidth]{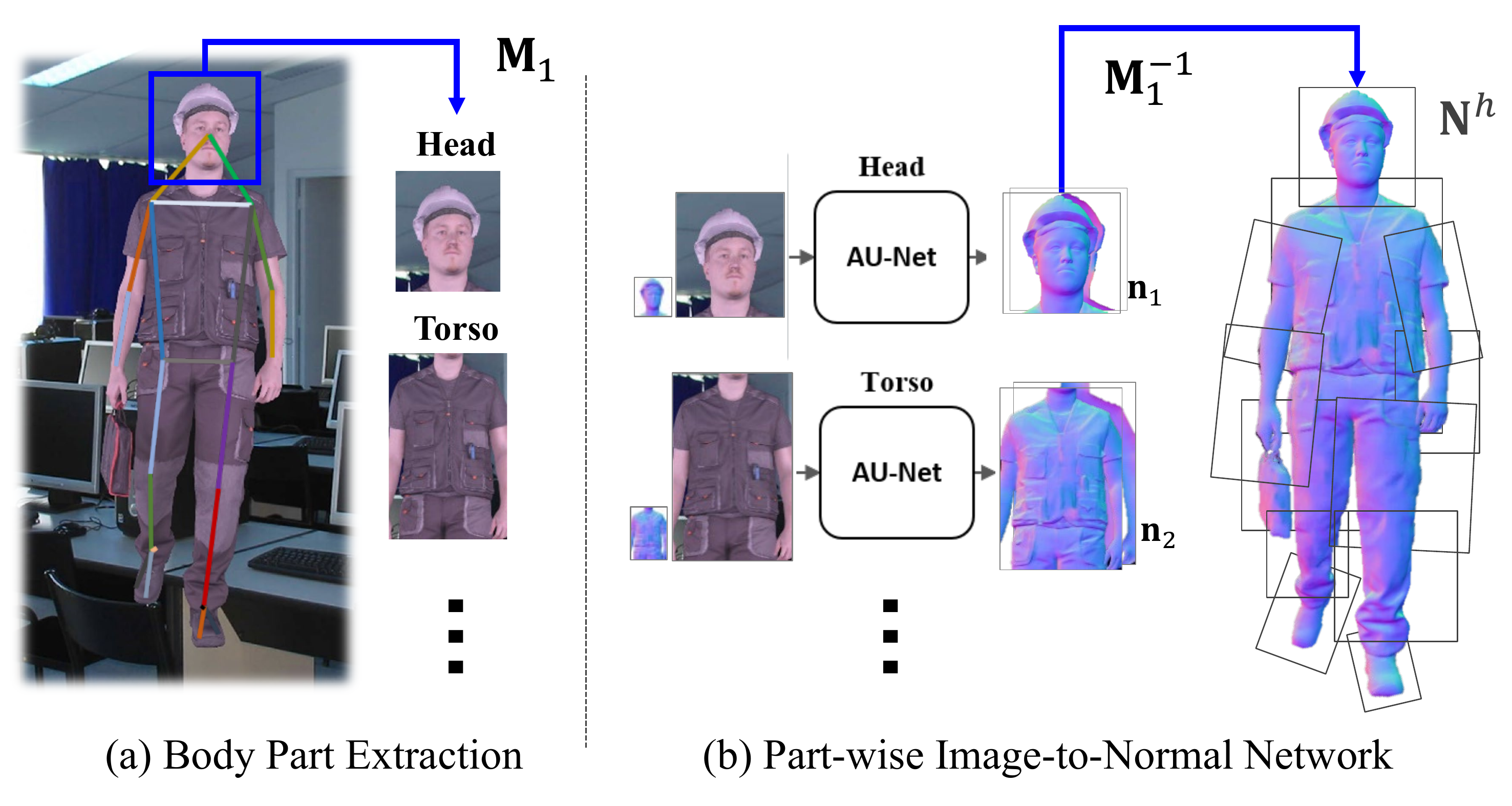}
\vspace{-20pt}
\caption{(a) Based on human joints, the body parts are categorized into five parts, \ie a head, a torso, arms, legs, and feet. Afterward, we transform them in the predefined sizes and positions and align the same body parts in the same direction to have a roughly similar structure. (b) We adopt the AU-Net \cite{Oktay_2018_AttentionUL} to generate front and back view part-wise normal maps $\mathbf{n}_i$. Then, we warp each normal map to the original position and scale with the transformation matrix $\mathbf{M}_i$, and then finally obtain the high-resolution normal maps $\mathbf{N}^h$.}
\vspace{-3mm}
\label{fig:body_part_extraction}
\end{figure}

\noindent\textbf{Body part extraction and alignment.}%
\textcolor{black}{Since the image-to-normal network takes aligned part images as input, we first crop image patches using 2D joints for the 12 body parts \ie, head, torso, upper arms, lower arms, upper legs, lower legs, and feet.}%
Let $\mathcal{P}=\{\mathbf{p}_i | 0\le i < 12 \}$ be the twelve patches cropped from image $I$ by using 2D joints $\mathcal{J}$, centered at 2 to 4 joints.  
For example, the upper arm region is extracted by using two joints, the shoulder and elbow, and the torso region is extracted by using two shoulder joints and two pelvis joints. %
Then, we warp image patches,  
\begin{equation}
    \begin{array}{l}
     \mathbf{\bar{p}}_i = \mathbf{M}_i \mathbf{p}_i,
    \end{array}
        \label{eq:body_part_extractor1}
\end{equation}
where $\mathbf{\bar{p}}_i\in\mathcal{\bar{P}}$ is the body part transformed by a similarity transformation $\mathbf{M}_i$.
An example of cropped images is illustrated in Fig.~\ref{fig:body_part_extraction} (a), where we feed $\mathcal{\bar{P}}$ to the normal-to-image network.
\textcolor{black}{An inverse similarity transformation $\mathbf{M}^{-1}_i$ is then used for merging the part-wise normal maps by inversely warping each patch.}
This pipeline is simple yet effective to train high-resolution images not only because it reduces the computational complexity but also because a less complicated network can be used for aligned part-wise depth prediction.

\noindent\textbf{Part-wise normal prediction.}
We generate the part-wise surface normals by using an image-to-normal network $G_{\mathbf{N},i}(\cdot)$ as below:
\begin{equation}
    \begin{array}{l}
    \textcolor{black}{\mathbf{\bar{n}}_i = G_{\mathbf{N},i}(\mathbf{\bar{p}}_i,  \mathbf{M}^{-1}_i\mathbf{N}^l),} 
    \end{array}
    \label{eq:normal_generator}
\end{equation}
where $\mathbf{\bar{n}}_i$ denotes a double-sided normal map for $i^\mathrm{th}$ aligned body part. The double-sided normal map indicates the stacked normal maps for front and rear views. 
\textcolor{black}{The $\mathbf{N}^l$ is used to separate the background and provide an initial geometry, which will be explained in Sec.~\ref{sec:low_resolution}.} 
Note that instead of training twelve independent networks, we train the five image-to-normal prediction networks for head, torso, arms, legs, and foot regions, respectively, as shown in Fig. \ref{fig:body_part_extraction} (b). The separation scheme enforces each network to focus on its unique structure and patterns \textcolor{black}{of the body parts} as well as to reduce the number of learnable parameters. In addition, we use the AU-Net~\cite{Oktay_2018_AttentionUL} as our backbone network to predict normal maps. 
Once normal maps are predicted for all body parts, we assemble them into the high-resolution normal maps $\mathbf{N}^h$ as follows:\vspace{-1mm}
\begin{equation}
    \begin{array}{ll}
    \mathbf{N}^h & = \sum\limits_{i=1}^{K}\left(\mathbf{W}_i\odot \mathbf{n}_i\right), 
    \label{eq:normal_map_merging}
    \end{array}
    \vspace{-1mm}
\end{equation}
where \textcolor{black}{$\mathbf{n}_i=\mathbf{M}^{-1}_i\mathbf{\bar{n}}_i$} is an inverse warped normal map of $\mathbf{\bar{n}}_i$ by applying $\mathbf{M}^{-1}_i$ and $\odot$ stands for an element-wise product. 
Since the multiple normal vectors can be projected onto the same pixel in the original image, we blend normal vectors when they are near the boundary region and projected multiple times. 
The blending weight $\mathbf{W}_i(x,y)$ is computed as, 
\begin{equation}
\begin{array}{ll}
\mathbf{W}_i(x,y) = \dfrac{ G(x,y) * \phi_i(x,y) }{ \sum_{i}{G(x,y) * \phi_i(x,y)} },\ \ \\ \\ 
\phi_i(x,y) = \left\{ \begin{array}{rcl}
1 & \mbox{if}~~\sum \mathbf{n}_i(x,y) \neq \mathbf{0^{\top}} \\ 
0 & \mbox{otherwise} 
\end{array}\right.,  
\end{array}
\label{eq:weight_map}
\end{equation}
where $(x,y)$ denotes a pixel location, $*$ stands for a convolution operator, and $\mathbf{0^{\top}}$ is the six dimensional zero vector. We compute a pixel weight $\mathbf{W}_i(x,y)$ w.r.t. the pixels having projected normal vectors which results in decreasing weights near the boundary regions conditioned by the size of a Gaussian kernel $G(\cdot, \cdot)$ with the boundary region $\mathbf{\sigma}=7px$. 

\begin{figure}[t]
\centering
\includegraphics[width=1.0\linewidth]{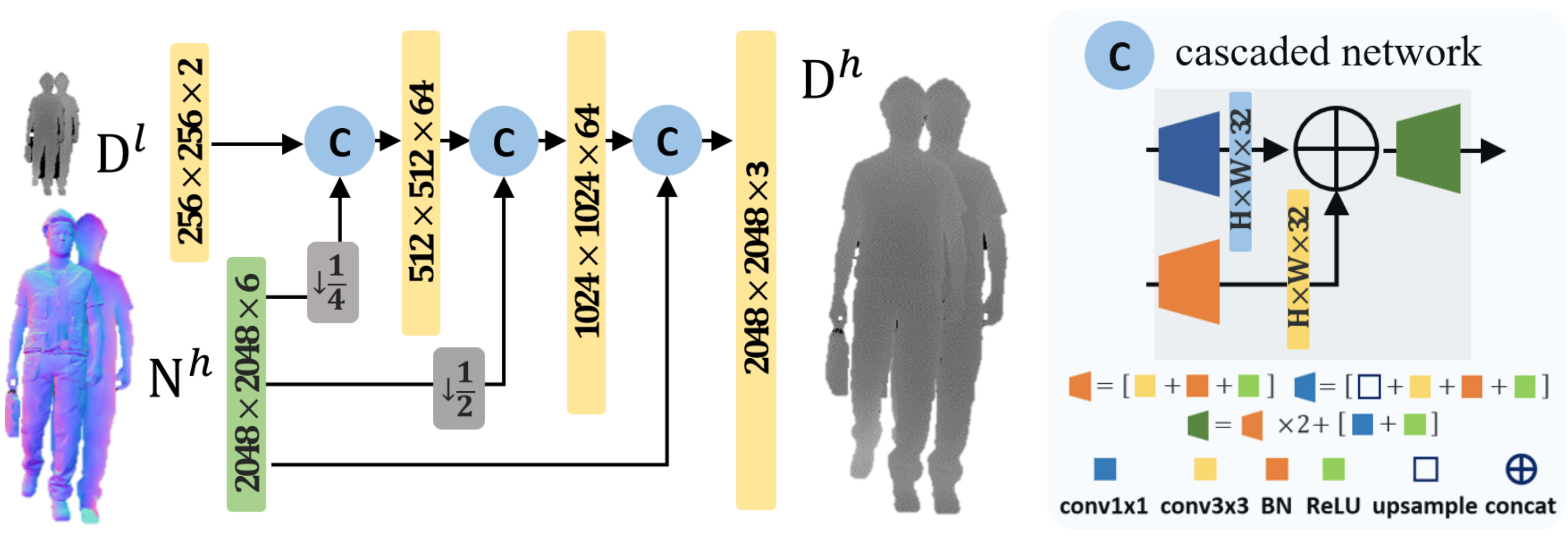}
\vspace{-20pt}
\caption{The high-resolution depth network takes the low-resolution depth and the high-resolution normal maps in \eqnref{eq:normal_map_merging} to render the high-resolution depth maps $\mathbf{D}^h$. All of normal maps and depth maps consist their front and back views.}
\vspace{-9pt}
\label{fig:rough_depth_estimator}
\end{figure}

\subsection{Low-resolution depth prediction network}
\label{sec:low_resolution} 

Rather than directly predicting the depth maps from the merged normal maps $\mathbf{N}^h$, we predict the coarse depth maps to guide predicting high-resolution depth maps.
\textcolor{black}{The low-resolution depth network sequentially infers coarse normal maps $\mathbf{N}^l$ and depth maps $\mathbf{D}^l$.}
Although it lacks geometric details, the smaller image size better captures the global geometric information.
In this work, we adopt the depth prediction network used in \cite{park2022monocular} because the dual-encoder AU-Net (D-AU-Net) selectively captures meaningful features from photometric and geometric information, therefore, different information can be complementary each other. On the other hand, the attention module makes the network converges quickly compared to the conventional U-Net structure. 
We first down-sample the input image $I$ to a low-resolution image $I^l$. The image-to-normal network in the D-AU-Net predicts double-sided coarse normal maps $\mathbf{N}^l$ from $\mathbf{I}^l$.
\textcolor{black}{Different from \cite{park2022monocular} which generates double-sided depth and normal maps, our low-resolution depth network $G_{\mathbf{D}^l(\cdot)}$ additionally yields a mask map $\mathbf{\hat{M}}^l$ to deal with the background. Finally, the low-resolution depth map $\mathbf{D}^l$ is computed by the element-wise product $\odot$ of $\mathbf{\hat{D}}^l$ and $\mathbf{\hat{M}}^l$.}
\begin{equation}
    \begin{array}{lll}
    \textcolor{black}{\mathbf{D}^l = \mathbf{\hat{D}}^l\odot\mathbf{\hat{M}}^l,} \\
    \textcolor{black}{\mathbf{\hat{D}}^l, \mathbf{\hat{M}}^l, \mathbf{N}^l = G_{\mathbf{D}}^l(I^l)}.
    \end{array}
    \label{eq:rough_depth_estimator}
\end{equation}
Here, we set the resolution of an input image to $256\times256$ which is smaller than the input by a factor of 8. 

\subsection{High-resolution depth prediction network}
We design the high-resolution depth network based on a shallow convolution network architecture that consists of three cascaded blocks as shown in Fig.~\ref{fig:rough_depth_estimator}.
The high-resolution depth network $G_{\mathbf{D}}^h(\cdot)$ is defined as below: 
\begin{equation}
    \begin{array}{lll}
    \textcolor{black}{\mathbf{D}^h = \mathbf{\hat{D}}^h\odot\mathbf{\hat{M}}^h,} \\
    \textcolor{black}{\mathbf{\hat{D}}^h, \mathbf{\hat{M}}^h = G_{\mathbf{D}}^h(\mathbf{N}^h, \mathbf{D}^l),}
    \end{array}
    \label{eq:high_depth_estimator}
\end{equation}
where it takes the low-resolution depth maps $\mathbf{D}^l$ and the high-resolution surface normal maps $\mathbf{N}^h$ as input to predict high-fidelity depth maps $\mathbf{D}^h$ \textcolor{black}{via the element-wise product $\odot$ of $\mathbf{\hat{D}}^h$ and $\mathbf{\hat{M}}^h$.}
The cascade block basically increase the resolution of input by the factor of 2, as illustrated in Fig.~\ref{fig:rough_depth_estimator}. Here, the first two blocks fuse depth and normal information to generate feature maps, and the last block generates double-sided depth map $\mathbf{D}^h$ from $\mathbf{N}^h$ and fused features. 
\textcolor{black}{Unlike the existing works that calculate the signed distance function through the implicit function of the MLP structure, we choose the CNN structure to encode the normal locality information. This contributes to forming a detailed geometry by exchanging information with adjacent pixels in the network structure.} 

\noindent\textbf{Mesh generation.} There are multiple ways to generate a 3D model from depth maps. In this work. we take a similar approach to \cite{Gabeur_2019_ICCV}. We convert the depth maps into 3D point clouds, and then compute a surface normal for each point from its neighboring points. Afterward, we run a screened Poisson surface construction~\cite{kazhdan2013screened} to obtain the smooth human mesh $\mathbf{V}$. 

\subsection{Training scheme and loss functions}

To train our network, we divide the training procedure into two phases while employing different loss functions.
\textcolor{black}{
In the \textit{phase1} of Fig.~\ref{fig:overall_framework}, we use both a $L_1$ loss and $L_{SSIM}$ loss to optimize the image-to-normal networks, $G_{\mathbf{N},i}(\cdot)$ and $G_{\mathbf{N}}^l(\cdot)$  as follows:
\begin{flalign}
    \begin{array}{ll}
    \!\!\!\!L_{\textit{phase1, N}}=\!\!\!\!\!&\alpha  L_{\text{\text{1}}}(\mathbf{N}^l_{GT}, \mathbf{N}^l) \!+\! \beta  L_{\text{SSIM}}(\mathbf{N}^l_{GT}, \mathbf{N}^l)\ \!+ \\
    &\!\!\!\!\!\!\!\!\!\sum_{i}(\alpha  L_{\text{\text{1}}}(\mathbf{\bar{n}}_{GT,i},~ \mathbf{\bar{n}}_i) \!+\! \beta  L_{\text{SSIM}}(\mathbf{\bar{n}}_{GT,i},~ \mathbf{\bar{n}}_i)),
    \end{array}
    \label{eq:losses_1N}
\end{flalign} 
where $L_{\text{SSIM}}$ denotes the SSIM loss to preserve high-frequency information in shape boundaries. 
The depth network $G_{\mathbf{D}}^l(\cdot)$ is then trained by minimizing a linear combination of a smooth-$L_1$ loss $L_{s1}$ and binary cross-entropy loss $L_{BCE}$ as below:
\begin{equation}
    \begin{array}{ll}
    \!\!\!\!L_{\textit{phase1, D}} =\!\!\!\!\!&\alpha  L_{\text{\emph{s}1}}(\mathbf{D}^l_{GT}, \mathbf{D}^l) \ \!+ \\
    &\beta  L_{\text{\emph{s}1}}(\mathbf{N}^l_{GT}, \mathbf{N}^l_D) \!+\! \gamma  L_{\textit{BCE}}(\mathbf{M}^l_{GT}, \mathbf{M}^l_D).
    \end{array}
    \label{eq:losses_1D}
\end{equation}
We also utilize a regularization term to maintain a consistency between the ground-truth normal maps and the normal maps $\mathbf{N}^l_D$ converted from the predicted depth maps $\mathbf{D}^l$, similar to ~\cite{Jafarian_2021_CVPR}. $L_{\textit{BCE}}$ is used to learn the foreground mask with the binary cross-entropy loss. Three hyper-parameters are empirically set to $\alpha$=0.85, $\beta$=0.15, and $\gamma$=0.15.}

For training the \textit{phase 2} in Fig.~\ref{fig:overall_framework}, we freeze the networks trained in the \textit{phase 1} and train for the high-resolution depth generator $G_{\mathbf{D}}^h(\cdot)$. For this, we use the same losses used in \eqnref{eq:losses_1D}:
\textcolor{black}{
\begin{equation}
    \begin{array}{ll}
    \!\!\!\!L_{\textit{phase2}} =\!\!\!\!\!& \alpha  L_{\text{\emph{s}\emph{l}1}}(\mathbf{D}^h_{GT}, \mathbf{D}^h) \ \!+ \\
    &\beta  L_{\text{\emph{s}\emph{l}1}}(\mathbf{N}^h_{GT}, \mathbf{N}^h_D) \!+\! \gamma  L_{\textit{BCE}}(\mathbf{M}^h_{GT}, \mathbf{M}^h_D),
    \end{array}
    \label{eq:losses_2}
\end{equation}
}
We also use the same hyper-parameters as in \eqnref{eq:losses_1D}.
Intuitively, fine-tuning the entire models can be a viable option. 

\section{Experimental Results}
\label{sec:exp}
For training, we use Adam optimizer~\cite{kingma2015adam} with $\beta_1=0.9$, $\beta_2=0.99$, and stochastic gradient descent with the warm restarts scheduler with $T_0=5$~\cite{loshchilov2016sgdr}. 
The learning rate and batch size are set to $0.0001$ and $2$, respectively. We set the kernel size to $15\times15$ and $\sigma$ to 5 for the Gaussian smoothing kernel ${G(x,y)}$.
We train our model for 30 epochs using the publicly available PyTorch framework, which takes about 3 days on a machine with four NVIDIA RTX A6000 GPUs.

\subsection{Training data generation}
Overall, we scanned 2,050 human models. We will release our human scan models to the public upon the acceptance of this paper. We also trained our network by using all the human models to see how far the proposed framework can improve the performance of human model generation. In addition, 368 high-quality human models~\cite{Renderpeople} are employed, denoted as the RenderPeople dataset, and \textcolor{black}{the THuman2.0~\cite{tao2021function4d} model was also used for this experiment.} 
We synthesized a background image from~\cite{quattoni2009recognizing} to the input image to increase the reconstruction efficiency for the real environment. 

To train our network, images and depth maps are generated for front and back views through perspective projection, and the normal maps are then obtained from the generated depth maps. First, we shift each human model to be located at the center of the rendered image using the centroid value. Then, we rotate the shifted model horizontally from -30 to 30 with an interval of 10 degrees. Here, the virtual camera position faces the subjects from the [0, 0, -1] position. The field of view is 50 degrees, and the image resolution is 2048$\times$2048.
In the rendering process, ambient and diffuse lighting is placed around the model to create a realistic training image. The ambient lighting applied to the object's center is fixed, the intensity and color of diffuse lighting are randomly changed, and 90 diffuse lights are placed around the object. As a result, shaded front images, back images, depth maps, and normal maps are generated for training.

In addition, we demonstrate the robustness of our network. To do this, we use two public datasets:
RenderPeople~\cite{Renderpeople} and THuman2.0 dataset~\cite{tao2021function4d}. 
We purchase 368 commercial subjects for RenderPeople dataset and split them whose train/test set ratio is about 9:1 (331/37).
For THuman2.0, the number of the train and test sets is 500 and 26, respectively, following the evaluation protocol in ICON~\cite{xiu2022icon} for fair comparison. 
As an ablation study, we demonstrate both the usefulness of our dataset and its synergy with our network by checking the performance improvement of our network with respect to the number of training images and its resolution. For the ablation study, we use up to 2,000 training subjects and 50 test subjects.

\begin{table}[t]
\setlength{\tabcolsep}{2pt}
\centering
\begin{center}
\renewcommand{\arraystretch}{1.2} 
    \resizebox{\linewidth}{!}{
	\begin{tabular}{c | c | c | c | c | c | c | c | c |c}
		\noalign{\hrule height 1.4pt}   
		 \multirow{2}{*}{\begin{tabular}{@{}c@{}}Method\end{tabular}} & \multicolumn{4}{c|}{RenderPeople} & \multicolumn{4}{c|}{THuman2.0} & \multirow{2}{*}{\begin{tabular}{@{}c@{}}\vspace{-2pt}Inference\\Time(s)$\downarrow$\end{tabular}} \\ 
		 \cline{2-9}
		  & {P2S$\downarrow$} & {Chamfer$\downarrow$} & {Normal$\downarrow$} &
		 {IoU$\uparrow$} & {P2S$\downarrow$} & {Chamfer$\downarrow$} & {Normal$\downarrow$} &
		 {IoU$\uparrow$} & \\ 
		 \hline
		 PIFu   & 1.96 & 2.10 & 8.11 & 64.1 & 3.10 & 3.02 & 8.52 & 59.5 & 4.16 \\
		 PIFuHD & 1.45 & 1.66 & 6.56 & 70.1 & 2.42 & 2.81 & 8.89 & 59.7 & 13.67\\
		 PaMIR  & 1.51 & 1.68 & 6.50 & 67.8 & 2.04 & 2.43 & 6.19 & 71.6 & 2.70\\
		 ICON   & \textbf{0.55} & \textbf{0.57} & 3.78 & \textbf{87.5} & \textbf{1.21} & \textbf{1.19} & 6.65 & \textbf{74.6} & 9.61 \\
		 \hline
		 \hline
		 2K2K   & 1.12 & 0.92 & \textbf{3.64} & 84.4 & 1.53 & 1.65 & \textbf{5.71} & 69.0 & \textbf{0.31}\\
         \noalign{\hrule height 1.4pt} 
	\end{tabular}
	}
\end{center}
\vspace{-18pt}
\caption{Single-view human reconstruction evaluation results on the RenderPeople and THuman2.0 datasets.} 
\vspace{-15pt}
\label{table:model_compare}
\end{table}

\subsection{Quantitative and qualitative evaluation}

We compare 2K2K with the state-of-the-art (SOTA) methods, including PIFu~\cite{Saito_2019_ICCV},  PIFuHD~\cite{Saito_2020_CVPR}, PaMIR~\cite{Zheng_2021_PaMIR} and \textcolor{black}{ICON~\cite{xiu2022icon} with human scanned data from RenderPeople~\cite{Renderpeople}, THuman2.0~\cite{tao2021function4d}.} For evaluation, we use popularly used metrics in the literature: average point-to-surface(P2S) distance, Chamfer distance, average surface normal error, and Intersection over Union(IoU). Metric units are centimeters for P2S and Chamfer, and radian for surface normal. \textcolor{black}{For the P2S distance and Chamfer distance metrics, RenderPeople used the original scale, and Thuman2.0 was evaluated by normalizing the mesh scale for fair evaluation because the scale is different for each subject.}

\begin{table}[t]
\setlength{\tabcolsep}{2pt}
\centering
\begin{center}
\renewcommand{\arraystretch}{1.2} 
    \scriptsize
    \begin{tabular}{c | c | c | c | c | c | c}
        \noalign{\hrule height 1pt} 
        \multirow{2}{*}{\begin{tabular}{@{}c@{}}Resolution\end{tabular}} & \multirow{2}{*}{\begin{tabular}{@{}c@{}}{P2S$\downarrow$}\end{tabular}} & \multirow{2}{*}{\begin{tabular}{@{}c@{}}{Chamfer$\downarrow$}\end{tabular}} & \multirow{2}{*}{\begin{tabular}{@{}c@{}}{Normal$\downarrow$}\end{tabular}} & \multirow{2}{*}{\begin{tabular}{@{}c@{}}\vspace{-2pt}Inference\\Time(s)$\downarrow$\end{tabular}} & \multicolumn{2}{c}{Memory(MB)} \\
        \cline{6-7}
         &  &  &  &  & Whole Img. & Part-wise\\
        \hline
        512$\times$512   & 1.17 & 1.20 & 4.72 & \textbf{0.043} & 6,565 & 5,735(87\%)\\
        1024$\times$1024 & 1.15 & 1.10 & 4.36 & 0.091 & 15,065 & 11,011(73\%)\\
        2048$\times$2048 & \textbf{0.96} & \textbf{1.01} & \textbf{3.69} & 0.308 & 47,195 & 30,413\textbf{(64\%)}\\
        \noalign{\hrule height 1pt} 
    \end{tabular}
\end{center}
\vspace{-17pt}
\caption{Performance\,evaluation\,\wrt\,the\,different\,resolution inputs. As the resolution of input becomes larger, our approach more effectively predicts the depth maps.\,Our method predicts high-quality depth maps at 20/10/3 fps for different resolutions of inputs.}
\vspace{-10pt}
\label{table:ablation_resolution}
\end{table}

\begin{figure}[t]
\centering
\includegraphics[width=1.0\linewidth]{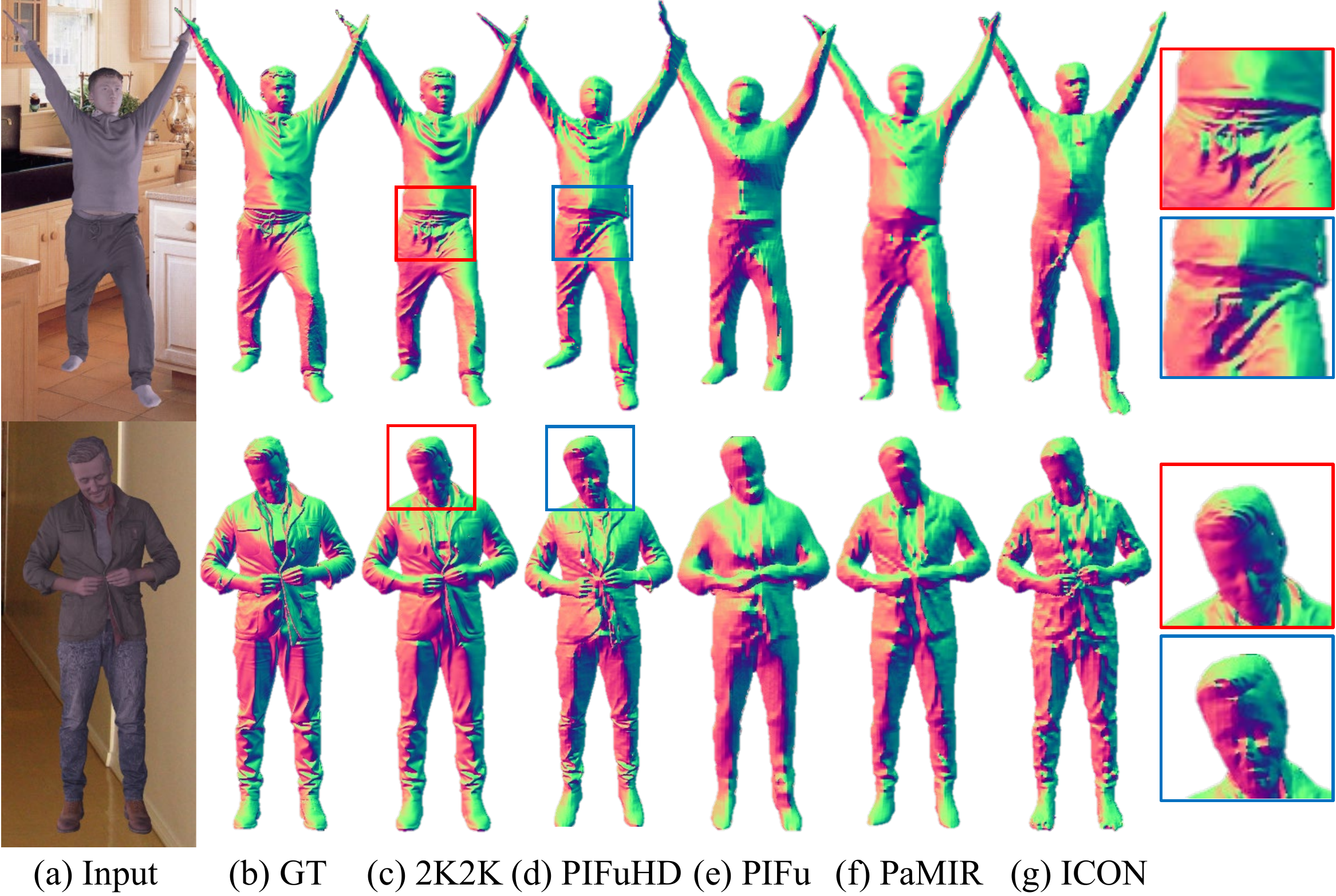}
\vspace{-17pt}
\caption{Qualitative results. 2K2K shows accurate results over (c)-(g). Particularly, our results show that the detailed shape of the human body are better expressed compared to other methods.}
\vspace{-15pt}
\label{fig:Comparison_other}
\end{figure}

\begin{figure*}[t]
\centering
\includegraphics[width=1.0\linewidth]{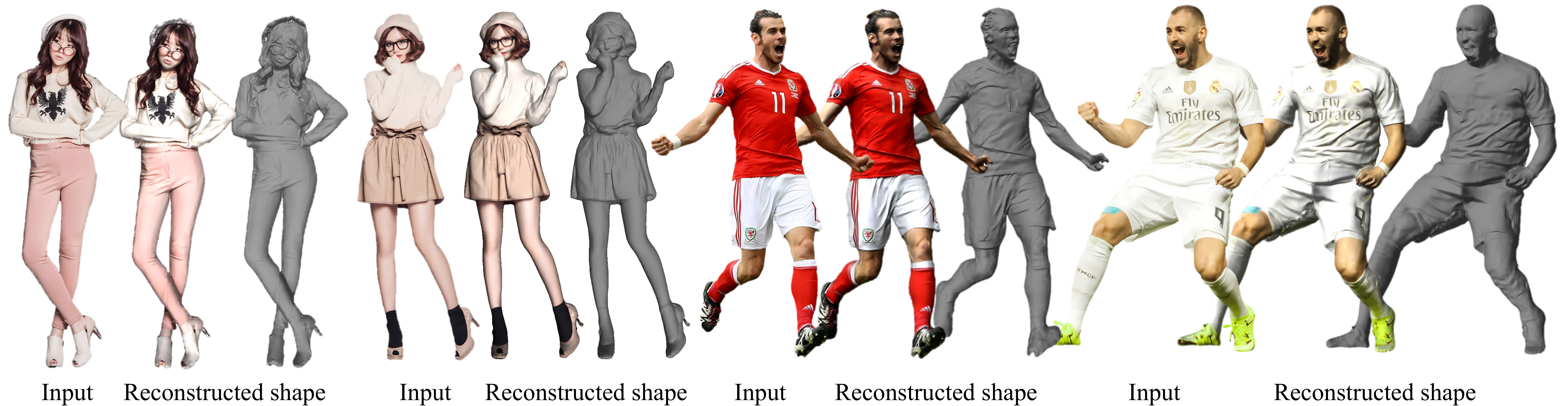}
\vspace{-6mm}
\caption{Single image 3D human reconstruction results in the wild. The images are downloaded in internet.}
\label{fig:real_image}
\vspace{-5mm}
\end{figure*}

\begin{figure}[t]
\centering
\includegraphics[width=1.0\linewidth]{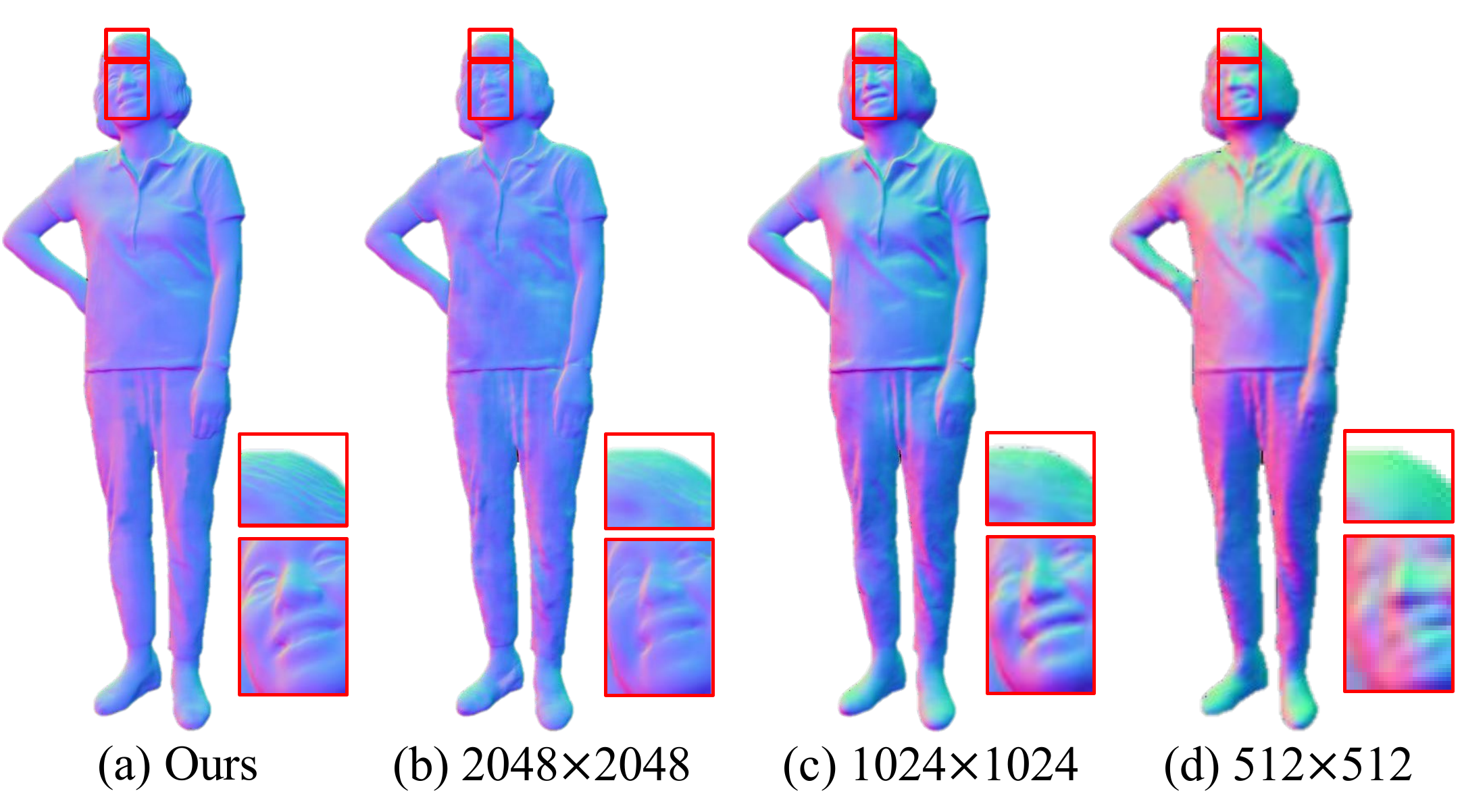}
\vspace{-22pt}
\caption{An example of our result. (b) is the part-ware result of the 2048 resolution image, and (c)~(e) are the results of direct prediction of normals in the 2048, 1024, and 512 resolution images.}
\vspace{-12pt}
\label{fig:Comparison_resolution}
\end{figure}

Table.~\ref{table:model_compare} shows the quantitative comparison results. 
Ours achieves the best results in almost measures. Although ICON shows promising result in terms of the P2S distance and Chamfer distance, we outperform the SOTA models in the surface normal metric, thanks to the strength of CNN architecture taking locality information to the end of the network.
We add its detailed explanation in supplementary.
For inference time, other implicit function methods require a huge time to compute the signed distance function, while since our method predicts the depth map in the end-to-end manner, the inference is done within 0.3 seconds even with 2K resolution images.
Fig.~\ref{fig:Comparison_other} compares the normal maps by using predicted human models from RenderPeople and THuman2.0 images. As shown, our method can recover details of facial regions compared to the other methods. 

\noindent\textbf{Effectiveness.}
We evaluated the memory footprint, \eg, required memory for a single batch training, at different resolutions. 
Table.~\ref{table:ablation_resolution} shows the actual memory usage during training and the perfomance changes. When we use 2K resolution images, we can obtain the impressive results as expected. In addition, our part-based approach enable our network to use the high-resolution images as input. That means that the proposed method is more efficient and powerful than those of the direct usage of whole images as input.

\noindent\textbf{Geometric detail.}
Table.~\ref{table:ablation_resolution} shows the performance according to the image resolutions in our dataset.
\textcolor{black}{The P2S and Chamfer metrics imply that there is a little performance gain when using higher-resolution images. However, we observe that the significant improvements exist in terms of the normal metric, which means high-frequency details are well preserved.}
In addition, Fig.~\ref{fig:Comparison_resolution} compares the different quality of outputs w.r.t the resolution changes. As expected, 1024$\times$1024 and 512$\times$512 results show blurry results because downsampling the image erases meaningful information. 
The most important comparison is in Fig.~\ref{fig:Comparison_resolution} (a) and (b) where the input resolution is same but the whole-body approach loses details compared to the part-wise approach. This indicates that the part-wise approach is not only efficient but also accurate, with the aid of the part-wise normal prediction scheme.

\begin{figure}
\vspace{6pt}
\centering
    \adjincludegraphics[width=.105\linewidth,trim={0 0 0 .05cm},clip]{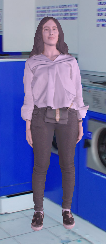}
    \animategraphics[loop,autoplay,width=.12\linewidth,trim = .8cm 0 .8cm 0]{10}{./images/ioys3/ioys3-}{0}{30}
    \adjincludegraphics[width=.105\linewidth,trim={0 0 0 .05cm},clip]{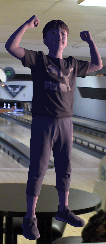}
    \animategraphics[loop,autoplay,width=.12\linewidth,trim = .8cm 0 .8cm 0]{10}{./images/ioys4/ioys4-}{0}{30}
    \adjincludegraphics[width=.105\linewidth,trim={0 0 0 .05cm},clip]{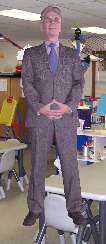}
    \animategraphics[loop,autoplay,width=.12\linewidth,trim = .8cm 0 .8cm 0]{10}{./images/rp2/rp2-}{0}{30}
    \adjincludegraphics[width=.105\linewidth,trim={0 0 0 .05cm},clip]{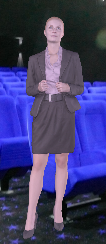}
    \animategraphics[loop,autoplay,width=.12\linewidth,trim = .8cm 0 .8cm 0]{10}{./images/rp1/rp1-}{5}{35}
\vspace{-5pt}
\caption{\!Animated\,full-viewing\,is\,available\,with\,Acrobat\,reader.}
\vspace{-10pt}
\label{fig:allaroundview_syn}
\end{figure}

We show animated reconstruction results while rotating the human model in Fig.~\ref{fig:allaroundview_syn}. 
We note that our method works well for non-front-facing humans because we rotate human models in the augmentation step. 
\textcolor{black}{In addition, Table.~\ref{table:ablation_dataset} shows the performance change according to the number of training datasets. The metric indicates that the overall performance is saturated when the number of the subjects is 2,000.}

\noindent\textbf{In-the-wild performance.}
Fig.~\ref{fig:real_image} shows the qualitative result made from internet photos. Although image and lighting information not included in the training set, our method shows high-fidelity reconstructions from in-the-wild images. 

\begin{table}[t]
\setlength{\tabcolsep}{12pt}
\centering
\vspace{7pt}
\begin{center}
\renewcommand{\arraystretch}{1.2} 
    \footnotesize
    \begin{tabular}{c | c | c | c  }
        \noalign{\hrule height 1pt}
        \# of scan models & {P2S$\downarrow$} & {Chamfer$\downarrow$} & {Normal$\downarrow$} \\
        \hline
        250  & 1.44 & 1.35 & 7.00 \\
        500  & 0.99 & 1.05 & 4.36 \\
        1,000 & 0.98 & 1.04 & 4.06 \\
        2,000 & \textbf{0.965} & \textbf{1.010} & \textbf{3.65} \\
        \noalign{\hrule height 1pt} 
    \end{tabular}
\end{center}
\vspace{-15pt}
\caption{A comparison of performance changes according to the number of training datasets. This table verifies that the number of scan models is an important factor.}
\vspace{-12pt}
\label{table:ablation_dataset}
\end{table}

\noindent\textbf{Limitations.}
Since we explicitly predict normal maps for each body part, our method do not take severe self-occlusion into account, \eg, when a lower arm is behind the back. We claim that this phenomenon is inherently ambiguous, possible remedies are either predicting semantics for occluding pixels or to employ human body prior~\cite{Pavlakos_2019_CVPR} to guide the depth prediction. We show several failure cases in the supplementary material due to the space limit.

\vspace{-2pt}
\section{Conclusion}
\vspace{-3pt}
\label{sec:conclusion}
We have proposed 2K2K, an effective framework for digitizing humans from high-resolution single images. To achieve this, we first built a large-scale human model dataset by scanning 2,050 human models, and used them to train our networks, consisting of part-wise normal prediction, low-resolution, and high-resolution depth prediction networks. To effectively handle the high-resolution input, we crop and weakly align each body part not only to handle pose variations but also to better recover fine details of the human body such as facial expressions. We demonstrated that the proposed method works effectively for the high-resolution images, for the various datasets.

\vspace{2mm}
\noindent\textbf{\small Acknowledgments.}
{\footnotesize This research was partially supported by the Ministry of Trade, Industry and Energy (MOTIE) and Korea Institute for Advancement of Technology (KIAT) through the International Cooperative R$\&$D program in part (P0019797), Institute of Information $\&$ communications Technology Planning $\&$ Evaluation (IITP) grant (No.2021-0-02068, Artificial Intelligence Innovation Hub / No.2022-0-00566. The development of object media technology based on multiple video sources), GIST-MIT Research Collaboration, “Practical Research and Development support program supervised by the GTI(GIST Technology Institute)" funded by the GIST in 2023, `Project for Science and Technology Opens the Future of the Region' program through the INNOPOLIS FOUNDATION funded by Ministry of Science and ICT, and the National Research Foundation of Korea (NRF) (No.2020R1C1C1012635) grant funded by the Korea government.}

\clearpage
{\small
\bibliographystyle{ieee_fullname}
\bibliography{refs}
}

\end{document}